\documentclass{bmvc2k}
\usepackage[ruled, lined, linesnumbered, commentsnumbered, longend]{algorithm2e}
\usepackage{algorithmic}
\usepackage{xcolor}
\definecolor{light-gray}{gray}{0.95}

\title{Likelihood-based Out-of-Distribution Detection with Denoising Diffusion Probabilistic Models}

\addauthor{Joseph Goodier}{jg852@bath.ac.uk}{1}
\addauthor{Neill D. F. Campbell}{ n.campbell@bath.ac.uk}{1}

\addinstitution{Centre for Accountable, Responsible and Transparent Artificial Intelligence (ART-AI), \\
University of Bath,\\
Bath, UK}

\runninghead{Goodier, Campbell}{Likelihood-based OOD Detection with DDPMs}

\begin{document}

\maketitle
\begin{abstract}

Out-of-Distribution detection between dataset pairs has been extensively explored with generative models. We show that likelihood-based Out-of-Distribution detection can be extended to diffusion models by leveraging the fact that they, like other likelihood-based generative models, are dramatically affected by the input sample complexity. Currently, all Out-of-Distribution detection methods with Diffusion Models are reconstruction-based. We propose a new likelihood ratio for Out-of-Distribution detection with Deep Denoising Diffusion Models, which we call the Complexity Corrected Likelihood Ratio. Our likelihood ratio is constructed using Evidence Lower-Bound evaluations from an individual model at various noising levels. We present results that are comparable to state-of-the-art Out-of-Distribution detection methods with generative models.
\end{abstract}
\vspace{-10pt}
\section{Introduction}
\label{sec:intro}

Out-of-Distribution (OOD) detection is a sub-class of uncertainty estimation that is a critical topic in the field of machine learning (ML). With the increasing complexity and scale of modern ML models, it has become increasingly important to understand the limitations and potential failure modes of these models. OOD detection refers to the ability of a model to identify inputs that are significantly different from those it was trained on, which is often achieved by looking at the uncertainty of a sample under a model. By detecting OOD inputs, ML models can avoid making incorrect or misleading predictions, which is especially important in high-stakes applications such as healthcare, finance, and autonomous driving. \

\begin{figure}[!h]
\hspace*{-0.5cm}
\centering
\includegraphics[width = 0.8\textwidth]{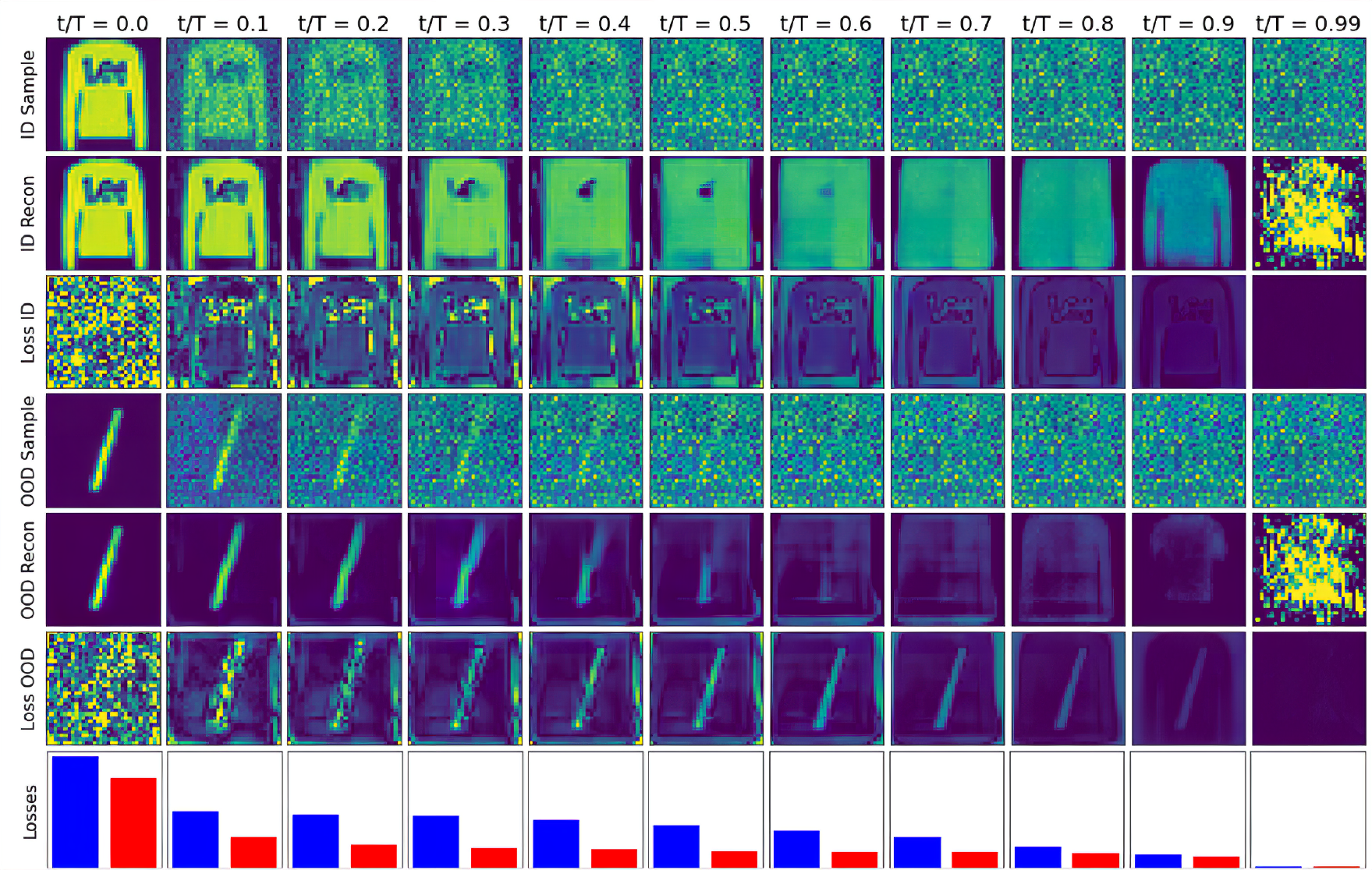}
\vspace*{0.25cm}
\caption{\small{Plots displaying how the noising level of an in-distribution (ID) (FashionMNIST, \textbf{Blue}) sample and OOD (MNIST, \textbf{Red}) sample affects the pixel-wise loss and reconstruction fidelity. The samples are reconstructed (\textbf{Second \& Fifth row}) in a single step, as opposed to iteratively. This plot demonstrates complexity bias as the more complex ID sample has a higher loss signal at lower noising levels than the less complex OOD sample (\textbf{Bottom row}), leading to a lower likelihood for the ID sample than the OOD sample. This plot also demonstrates that low noising levels, which are most corrupted by image complexity, are where low-level image features emerge.}}
\end{figure}

Generative models are particularly well suited for OOD detection because they are trained to model the distribution of the data, 
as opposed to the supervised learning approach that learns to map samples to labels. Likelihood estimations can be used to assess how likely a given sample is, under the learned model. The use of the likelihood for OOD detection has been applied to Variational AutoEncoders (VAEs) \cite{kingma2013auto}, flow-based models like GLOW \cite{kingma2018glow} and auto-regressive-based models like PixelCNN++ \cite{salimans2017pixelcnn++}. \

Denoising Diffusion Probabilistic models (DDPMs) \cite{ho2020denoising} provide a new addition to the landscape of generative models. DDPMs are generative models that add noise to an input sample and remove it using a noise prediction network.
The amount of noise added is learned during training, allowing pure noise inputs to be denoised and samples to be generated based on the learned distribution. All current methods that use DDPMs for OOD detection do so with reconstruction-based methods. Similar to VAEs, DDPMs are trained using the Evidence Lower-Bound (ELBO) objective as an estimate of the marginal log-likelihood of the model. It follows that DDPMs could also be used for likelihood-based OOD detection. \
\textbf{In this paper} we present :
\begin{enumerate}
    \item Evidence that input sample complexity dramatically affects the ELBO contributions from low noising levels in DDPMs, as is seen with other generative models.
    
    \item A likelihood-based OOD detection method using DDPMs. We use a likelihood ratio that is calculated using ELBO evaluations from low noise levels over the total ELBO from all noise levels. We define it to be the \textbf{Complexity Corrected Likelihood Ratio} (CCLR).

\end{enumerate}

\subsection{Likelihood-based OOD detection}

Bishop \textit{et al.} \cite{bishop1994novelty} proposed the use of model likelihood for OOD detection, where the author suggested it could be viewed as a probability under the model. Here, a sample would be assigned as OOD given a one-tailed test using the likelihood calculated by the model. In an ideal setup, a model would assign a high likelihood to in-distribution (ID) samples and a low likelihood to OOD samples. Recently, a number of contemporaneous publications challenged this assumption \cite{choi2018waic, hendrycks2018deep, nalisnick2018deep}. Specifically relating to generative models, Nalisnick \textit{et al.} \cite{nalisnick2018deep} and Choi \textit{et al.} \cite{choi2018waic} claim that generative models, when trained on an ID dataset and treat another dataset as OOD, similar to the above paradigm, assign higher likelihood scores to some OOD data. In light of this result, the community has released a number of studies that investigate and seek to redeem the likelihood for OOD detection. \

Ren \textit{et al.} \cite{ren2019likelihood} claim that the reason these likelihood-based methods fail is that the likelihood is dramatically affected by population-level background statistics that are not relevant to the specific details of the ID data. To remedy this, the authors propose a likelihood ratio score between the likelihood of a sample under the model and a background model that is trained on random perturbations of the input data. In a similar vein, Serra \textit{et al.} \cite{serra2019input} claim that the corruption of the likelihood score is related to the complexity of the test sample. The authors propose a likelihood ratio, based on an offset to the likelihood that is calculated with a loss-less compression of the sample. Building on the likelihood ratio, Xiao \textit{et al.}  \cite{xiao2020likelihood} propose another likelihood-based OOD score with a single model, called Likelihood Regret. The score is the difference in the likelihood of a sample under a model trained on the ID data, and the likelihood of the sample under the same model when it is over-fit to the sample. The greater the change in likelihood, the greater the likelihood regret score. \

Havtorn \textit{et al.} \cite{havtorn2021hierarchical} claims that the low-level features learned by VAEs generalise well across datasets and dominate the likelihood scores. They propose using a hierarchical VAE, where features at different levels in the input data are learned at the different levels in latent representations. Ran \textit{et al.} \cite{ran2022detecting}  propose a method using noise contrastive priors that they call Improved Noise Contrastive Priors (INCP), where they leverage synthetic OOD data created from ID samples to train a VAE. This contrastive prior can then be used to calculate an ELBO ratio, which is used as an OOD detection metric, with very high levels of success. Zhang \textit{et al.} \cite{zhang2021out} demonstrate that OOD generalisation ability depends on non-local features, which are defined in opposition to local or low-level features. The authors estimate a non-local likelihood score by calculating a likelihood ratio of two auto-regressive models; one trained on local features and the other trained on global features. Other methods involve using the Watanabe-Akaike Information Criterion (WAIC) for an ensemble of models as an OOD detection score \cite{choi2018waic}, whilst Morningstar \textit{et al.} \cite{morningstar2021density} use Density-of-States to assess the typicality of model statistics as an OOD detection score. \

The common thread across these contributions is that the likelihood of a sample under a generative model is corrupted by low-level or local features, that generalise well between datasets. These features contribute in an outsized amount to the complexity of an image, where we define image complexity as \textit{a measure of the degrees of freedom required to represent the images in the dataset}. The community has shown that what's actually important for OOD detection are the high-level, non-local semantics of an image that are unique to the dataset. The likelihood in generative models captures both of these, and the low-level image complexity needs to be corrected for in some way to produce a robust, likelihood-based OOD detection method with generative models.
\vspace{-10pt}
\subsection{Deep Denoising Diffusion Models}

DDPMs \cite{sohl2015deep, ho2020denoising} have emerged as a new class of generative models that have proven to be powerful and scaleable in many contexts. They are similar to normalising flows in that they map the input data to latent variables of the same dimension. They also resemble VAEs, as they estimate the likelihood through the ELBO objective \cite{prince2023understanding}. Unlike VAEs, instead of a learned encoder, the noise is added to the data in a defined schedule. The learning is done in the inverse process by a noise prediction network, that sequentially ``denoises'' the data. This inverse process is then used to produce samples from pure noise. We refer the reader to Yang \textit{et al.} \cite{yang2022diffusion} for a thorough review of various other diffusion model implementations and applications. \
 
\subsection{Diffusion-based OOD Detection}
A number of OOD detection methods that use DDPMs have been proposed. However, these methods all rely on reconstruction-based scores. The intuition of a reconstruction-based approach is that generative models should be able to faithfully reconstruct ID samples, and unfaithfully reconstruct OOD samples. The difference between the original sample and its reconstruction can be used as an OOD score. \

Graham \textit{et al.} \cite{graham2022denoising} use diffusion models to sequentially reconstruct samples for multiple noising levels. They use a combination of MSE and the LPIPS metric, which uses the distance between the deep features of the sample and reconstruction \cite{zhang2018unreasonable}, as an OOD metric. Liu \textit{et al.} \cite{liu2023unsupervised} employ diffusion models for inpainting occluded samples. They also use the LPIPS metric, where they calculate the median distance between the deep features of the inpainted reconstruction and the original, unoccluded sample as an OOD score. In a related approach, Liu \textit{et al.} \cite{liuout} use a combination of a discriminator and DDPMs as an asymmetric interpolation method for OOD detection. \

As previously stated, similar to VAEs, DDPMs are trained using the ELBO objective. It follows that DDPMs could also be used for likelihood-based OOD detection. These methods, when applied to other generative models, have historically demonstrated strong performance in OOD detection (See Table. 2) and have been explored more extensively than reconstruction-based methods. We posit that DDPMs, by extension, should also be able to exhibit a strong performance with likelihood-based OOD detection, given the current understanding of the considerations required to make likelihood-based OOD detection robust (See Fig. 1). 
\vspace{-10pt}
\section{Methodology}
\subsection{Model}
The model implementation used in this work is based on the discrete-time DDPM first introduced by Ho \textit{et al.} \cite{ho2020denoising}. Discrete-time DDPMs work by contaminating a sample, $\mathbf{x}_{0}$, with a distribution of noise defined by a time variable $t$, where  $0 \leq t \leq T$. At $T$, the sample has been degraded to isotropic noise. The noise distribution added at each $t$ is defined by a noising schedule, $\beta_{t}$, that produces noised samples $\mathbf{x}_{t}$.  We employ the cosine noising schedule introduced by Nichol \textit{et al.} \cite{nichol2021improved},
\begin{equation}
    \bar{\alpha}_{t} = \frac{f(t)}{f(0)}, f(t) = cos\left ( \frac{t/T + s}{1 + s}*\frac{\pi }{2}\right )^{2},
\end{equation}
where $\alpha_{t} = 1 - \beta_{t}$, $\bar{\alpha}_{t} = \prod_{s = 0}^{t}\alpha_{s}$ and $s>0$ is an offset to ensure that $\beta_{t}$ is not too small at $t = 0$. This schedule was chosen because Nichol \textit{et al.} \cite{nichol2021improved} showed that a cosine schedule was shown to distribute the information that emerges from DDPMs when sampling more smoothly across $t$-values than a linear schedule. The noising process can then be defined as, 
\begin{equation}
    q(\mathbf{x}_{t}|\mathbf{x}_{0}) = \mathcal{N}(\mathbf{x}_{t}|\sqrt{\bar{\alpha}_{t}}\mathbf{x}_{0}, (1 - \bar{\alpha})\mathbf{I}). 
\end{equation} 
For the reverse process, a network learns the noise distribution added for each timestep, which is then removed from the sample. The reverse process can be defined as, 
\begin{equation}
    p_{\theta}(\mathbf{x}_{t - 1}|\mathbf{x}_{t}) = \mathcal{N}(\mathbf{x}_{t - 1}|\mathbf{\mu}_{\theta}(x_{t},t), \Sigma_{\theta}(x_{t},t)),
\end{equation} 

where both $\mathbf{\mu}_{\theta}(x_{t},t)$ and $\Sigma_{\theta}(x_{t},t)$ are learned by the network, which is parameterised by $\theta$. During the learning process, a DDPM optimises the ELBO objective, which is the variational bound on the marginal log-likelihood \cite{ho2020denoising}. In practice, the $\Sigma_{\theta}(x_{t},t)$ term in Eq. 3 is fixed as a function of $t$ i.e. the variance is now $\Sigma(t)$; which then allows the model to predict solely the noise term for each timestep. It follows that the ELBO reduces down to a sum of the L2 distance between the predicted and actual noise for each timestep,
\begin{equation}
    \mathcal{L}_{\theta} = \frac{1}{T}\sum_{t = 0}^{T}\left [\left \|\hat{\mathbf{\epsilon}}_{\theta,t} - \mathbf{\epsilon}\right \|^{2}\right ],
\end{equation}
where $\epsilon$ is the sample of isotropic Gaussian noise, and $\hat{\mathbf{\epsilon}}_{\theta,t}$ is the approximation of the noise from the network. For a full derivation of the ELBO for a DDPM and how it reduces to Eq. 4, please see Prince, Ch. 18 \cite{prince2023understanding}. 

\subsection{Complexity Bias in DDPMs}
\begin{figure}[!h]
\centering
\includegraphics[width = 0.7\textwidth]{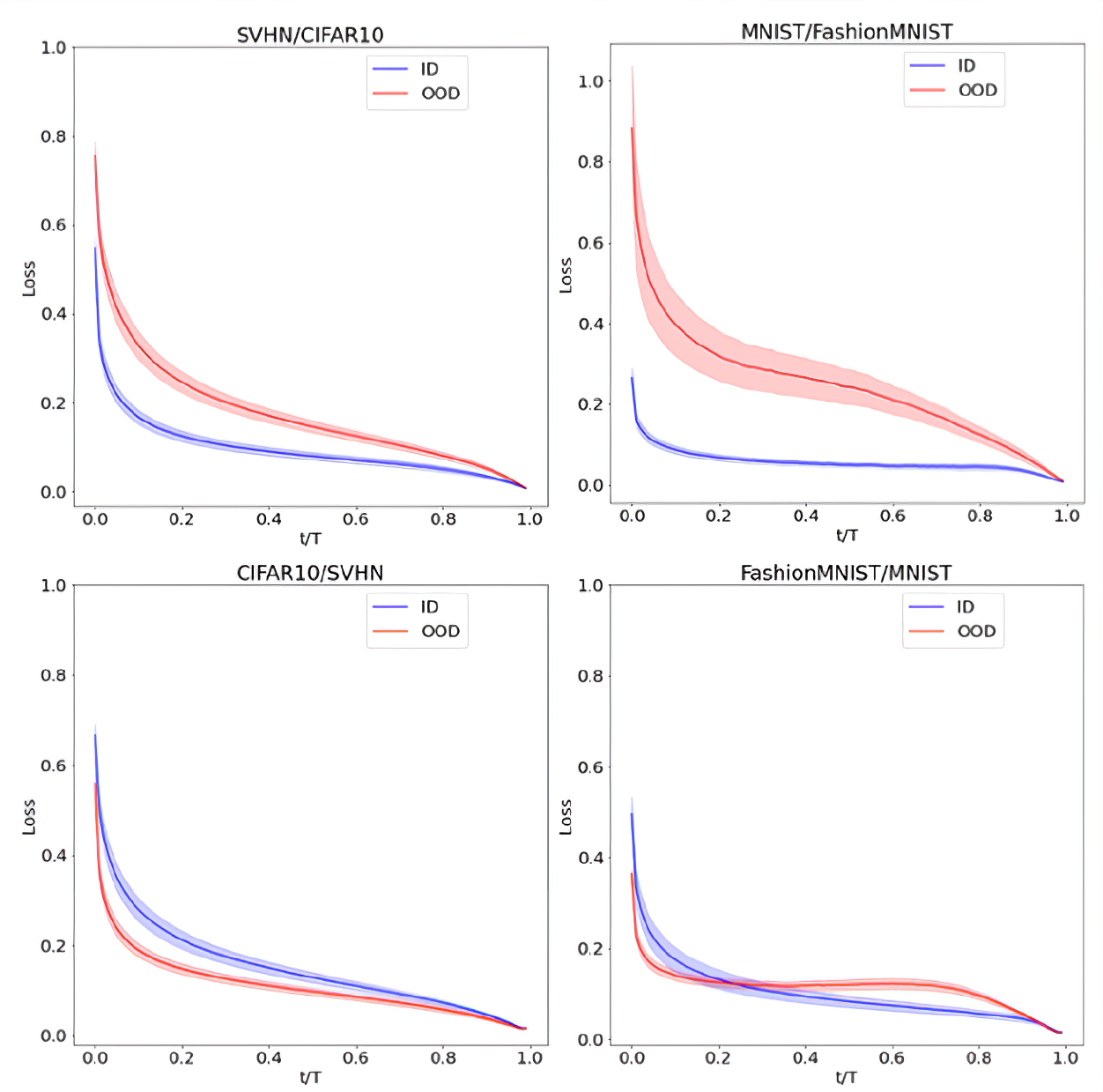}
\caption{\small{ A plot showing the average loss as a function of $t/T$ for dataset pairs. The top two plots demonstrate the expected behaviour that the OOD datasets have a higher average loss and therefore a lower log-likelihood under the model for SVHN/CIFAR10 (\textbf{Top Left}) and MNIST/FashionMNIST (\textbf{Top Right}). However, when the more complex dataset is ID, this expected behaviour no longer holds and there is a higher loss for ID dataset across $t/T$ for CIFAR10/SVHN (\textbf{Bottom Left}) and for low $t/T$ for FashionMNIST/MNIST (\textbf{Bottom Right}), and therefore a lower log-likelihood under the model.} }
\end{figure} 
It is well-established that generative models can assign a higher likelihood to some OOD samples than to ID samples \cite{nalisnick2018deep,choi2018waic,havtorn2021hierarchical}. Serra \textit{et al.} \cite{serra2019input} relate this to the idea that ``generative models exhibit a strong bias towards the complexity of the corresponding inputs''. They provide evidence that there is a negative correlation between the image complexity of the input samples and the log-likelihood scores. The qualitative behaviour of log-likelihood is the same as the likelihood i.e. OOD samples should have a low log-likelihood under the model and vice versa for ID samples. \

When training a vanilla DDPM, you are noising the sample to some $t$-value, predicting the noise that was added to the sample and taking a difference between actual noise and predicted noise as your loss. For low $t$-values, the noise added to a sample is very subtle (See Fig. 1). Therefore, for less complex images, it is easier to predict the noise and therefore the loss should be lower, even if these low-complexity samples are OOD. It follows then that the ELBO contributions from low $t$-values should also be affected by the complexity of the image statistics by an outsized amount through this process. In practice, when you are optimising the ELBO, you are minimising the upper bound on the negative marginal log-likelihood. Therefore, a \textit{lower} loss implies a \textit{higher} log-likelihood estimate.\

It follows that an OOD sample with lower complexity can have a higher marginal log-likelihood estimate than an ID sample with higher complexity. This effect can be seen in DDPMs in Fig. 2, where we compare the average loss contributions across all $t$-values for RGB and single-channel dataset pairs of differing complexity. For the RGB dataset pairs (\textbf{Left column}) the lower loss scores for the less visually complex SVHN, when both ID and OOD, imply a higher marginal log-likelihood estimate and vice versa for CIFAR10. However, the difference is less when SVHN is the OOD class. For the single-channel image pairs (\textbf{Right column}), this difference is very prominent in the MNIST/FashionMNIST where the low-complexity MNIST is also the ID dataset. For the FashionMNIST/MNIST dataset pair, for low $t$-values, the lower complexity MNIST dataset has lower loss values. But as the $t$-values increase, the ID class displays lower loss values and, therefore, a higher marginal log-likelihood estimate. This demonstrates that image complexity corrupts the ELBO contributions from low $t$-values, and needs to be corrected for to produce a robust likelihood-based OOD detection method using DDPMs.
\vspace{-5pt}
\subsection{Likelihood Ratio for DDPMs}
From Eq. 4, we define a decomposed ELBO for DDPMs as, 
\begin{equation}
    \mathcal{L}^{<k}_{\theta} = \frac{1}{k}\sum_{t = 0}^{k} w(t) \left[\left \|\hat{\mathbf{\epsilon}}_{\theta,t} - \mathbf{\epsilon}\right \|_{1}\right ],
\end{equation}
where $k$ is a noising threshold defined by $0 < k < T$. $\mathcal{L}^{<k}_{\theta}$ defines the ELBO for samples noised up to the threshold $k$. For a well-chosen value of $k$, $\mathcal{L}^{<k}_{\theta}$ should capture low $t$-value loss contributions that are affected by sample complexity. \

During training and inference, we use an L1 rather than an L2 loss. The main reason for this change relates to the findings of Saharia \textit{et al.} \cite{saharia2022palette} that the use of L1 reduces sample diversity, which in turn reduces the probability of hallucinations in diffusion models. This is important for OOD detection, as a tightly learned distribution over the training data is critical for the ability of the method to identify OOD samples. The L1 loss has also been shown to lead to increased training stability for DDPMs \cite{chen2020wavegrad, saharia2022image}. \

We also include a linear importance weight $w(t) = 0.5 + t/T$. Various DDPM implementations in the literature have been shown to be special cases of weighting the loss, and have been shown to improve the performance of DDPMs in terms of image generation quality \cite{kingma2021variational}. Kingma \textit{et al.} \cite{kingma2023understanding} show that if a weighting term is monotonic with $t$, then ``the weighted loss corresponds to maximizing the ELBO of noise-perturbed data''. Our coefficient $w(t)$ increases the importance of the L1 contributions added at higher $t$-values, to reduce the influence on the ELBO of the complexity bias at low $t$-value contributions. This term also has the property that, $\frac{1}{T}\int_{t=0}^{T}w(t)dt = 1$. \

The $\mathcal{L}^{<k}_{\theta}$ can be interpreted as the background model that captures background statistics from the original likelihood ratio for OOD work by Ren \textit{et al.} \cite{ren2019likelihood}, but for DDPMs and using only one model. It's worth noting here that if $k = T$, then $\mathcal{L}^{<T}_{\theta}$ becomes the full DDPM ELBO term  $\mathcal{L}_{\theta}$. The only difference is the importance weights $w(t)$. If $k$ is well chosen, it is possible to construct a likelihood ratio that corrects for the influence of complexity bias that affects the ELBO score as an OOD estimate. We define this ratio to be the CCLR, 
\begin{equation}
     CCLR_{k/T} = \mathcal{L}^{<k}_{\theta} - \mathcal{L}^{<T}_{\theta}.
\end{equation}
The ordering is due to the fact that $\mathcal{L}^{<k}_{\theta}$ and $\mathcal{L}^{<T}_{\theta}$ are ELBO estimates, which implies they are bounds on the \textit{negative} marginal log-likelihood. Therefore, they are multiplied by $-1$ so that the likelihood ratio has the property that an OOD sample should have a low likelihood ratio under this model. CCLR is also a difference in the log-likelihood space, which implies a ratio in the likelihood space. See Ablation in Appendix A for justification of algorithmic implementation choices.
\vspace{-13pt}
\subsection{OOD Detection Algorithm}
In order to optimise CCLR to perform OOD detection with a DDPM, we have to take a number of model-specific steps to enable an inference estimate. These steps are to ensure that both $\mathcal{L}^{<k}_{\theta}$ and $\mathcal{L}^{<T}_{\theta}$ are calculated from the same number of sample estimates. Firstly, each sample is expanded and passed through the model as a batch of size $N_{bs}$. This is to ensure that we get sufficient coverage of $t$-values to marginalise over, as the batch size of the samples, $\textbf{xs}$, must match the batch size of the $t$-values, $\textbf{ts}$, when passed through the DDPM. However, the $\mathcal{L}^{<k}_{\theta}$ term is calculated from $T/k$ times fewer samples than the full $\mathcal{L}^{<T}_{\theta}$ term. To account for this, we pass the batch through the model $n$ number of times, where $n = T/k$, combine the L1 losses across the $t$ dimension and then calculate the $\mathcal{L}^{<k}_{\theta}$ term. This corrects for the difference of samples across $t$ with samples across noise. The full algorithm for OOD detection is as follows:

\begin{algorithm}
    \algsetup{linenosize=\small}
    \footnotesize
    \caption{\footnotesize{Complexity Corrected Likelihood Ratio (CCLR)}}\label{LR}
    \begin{flushleft}
        \textbf{INPUT:} Data sample $x_{0}$, noising threshold $k$, inference batch-size $N_{bs}$,  diffusion model $\theta$\\
        \textbf{OUTPUT:} $CCLR_{k/T}$
    \end{flushleft}
  \begin{algorithmic}[1] 
  \item $ \mathbf{ts} = $ \texttt{linspace}$(0,T,N_{bs})$
  \item $\mathbf{w} = $ \texttt{linspace}$(0.5,1.5,N_{bs})$
  \item $\mathbf{xs} =$ \texttt{expand}$(N_{bs}, \mathbf{x}_{0})$
  \item $\mathcal{L}^{<T}_{\theta}$ = \texttt{mean}$(\mathbf{w}*$\texttt{diffuse}$(\mathbf{xs}, \mathbf{ts}))$
  
  \item $n = T/k$
  \item $\mathbf{ls} = [ \ ]$
    \FOR{$\_$ in \texttt{range}$(n)$} 
    \item $\mathbf{ls}.\texttt{append}(\texttt{diffuse}(\mathbf{xs}, \mathbf{ts}$))
    \ENDFOR   
    
  \item $\mathcal{L}^{<k}_{\theta} = \texttt{mean}\mathbf{(w}*$\texttt{logsumexp}$(\mathbf{ls})[:k])$
  \item $CCLR_{k/T} = \mathcal{L}^{<k}_{\theta} - \mathcal{L}^{<T}_{\theta}$
\end{algorithmic}
\end{algorithm}
\vspace{-15pt}
\section{Experiments \& Results}
\vspace{-10pt}
\subsection{Model Implementation}
The backbone DDPM code is based on the discrete-time DDPM from the library $\texttt{denoising}$
$\texttt{-diffusion-pytorch} $\cite{Wang2020}. All models were instantiated with $T$=1000 timesteps, trained for 100 epochs and with a learning rate of $2.0 \times 10^{-5}$. A cosine noising schedule was used for all trials, following Nichol \textit{et al.} \cite{nichol2021improved}. We use the L1 norm as the training loss for all objectives. The noise prediction network was instantiated with 64 feature maps for the initial block. There were 4 blocks, each with multipliers for the feature maps of (1,2,4,8), which result in map resolutions of 32$\times$32, 16$\times$16, 8$\times$8 and 4$\times$4 respectively.
\vspace{-10pt}
\subsection{Experiments}
For single-channel dataset pairs, we train on FashionMNIST (Xiao \textit{et al.} \cite{xiao2017fashion}) and treat the test set as ID. The OOD test set will be MNIST (Lecun \textit{et al.} \cite{lecun1998gradient}). For RGB dataset pairs, we train on CIFAR10 (Krizhevsky \cite{krizhevsky2009learning}) and treat the test set as ID. The OOD test set will be SVHN (Netzer \textit{et al.} \cite{netzer2011reading}). All images are resized to $32\times32$. For all dataset pairs, we randomly sample 1000 images from the ID test set and the OOD test set and asses the method's ability to distinguish whether a sample is ID or OOD, on a sample-by-sample basis. We expand each sample to an inference batch size of $N_{bs} = 100$. We use  the Area-Under-Receiver-Operating Characteristics (AUROC) as the baseline metric to evaluate the performance of OOD detection methods \cite{nalisnick2018deep, choi2018waic, ren2019likelihood, serra2019input, xiao2020likelihood, havtorn2021hierarchical, zhang2021out}. 
\vspace{-10pt}
\subsection{Results}
\begin{table}[!h]
\centering
\scalebox{0.75}{
\begin{tabular}{lllllll}
\hline
\textbf{Dataset Pair}           & \multicolumn{6}{c}{\textbf{AUROC scores}}                                                     \\
                       & $\mathcal{L}^{<T}_{\theta}$ & $CCLR_{1/2}$ & $CCLR_{1/3}$ & $CCLR_{1/5}$ & $CCLR_{1/10}$ & $CCLR_{1/20}$ \\ \hline
FashionMNIST/MNIST & 0.784                  & 0.985     & \textbf{1.000}      & \textbf{1.000}      & \textbf{1.000}       & \textbf{1.000}       \\
CIFAR10/SVHN        & 0.941                & \textbf{0.964}     & 0.958     & 0.955     & 0.960       & 0.947      
\end{tabular}}
\vspace{0.2cm}
\caption{\small{AUROC scores for FashionMNIST/MNIST and CIFAR10/SVHN dataset pairs. Results are reported for the whole ELBO objective as an OOD score, as well as likelihood ratios for various $k/T$-values.}}
\end{table}

\begin{figure}[!h]
\centering
\includegraphics[width = 0.9\textwidth]{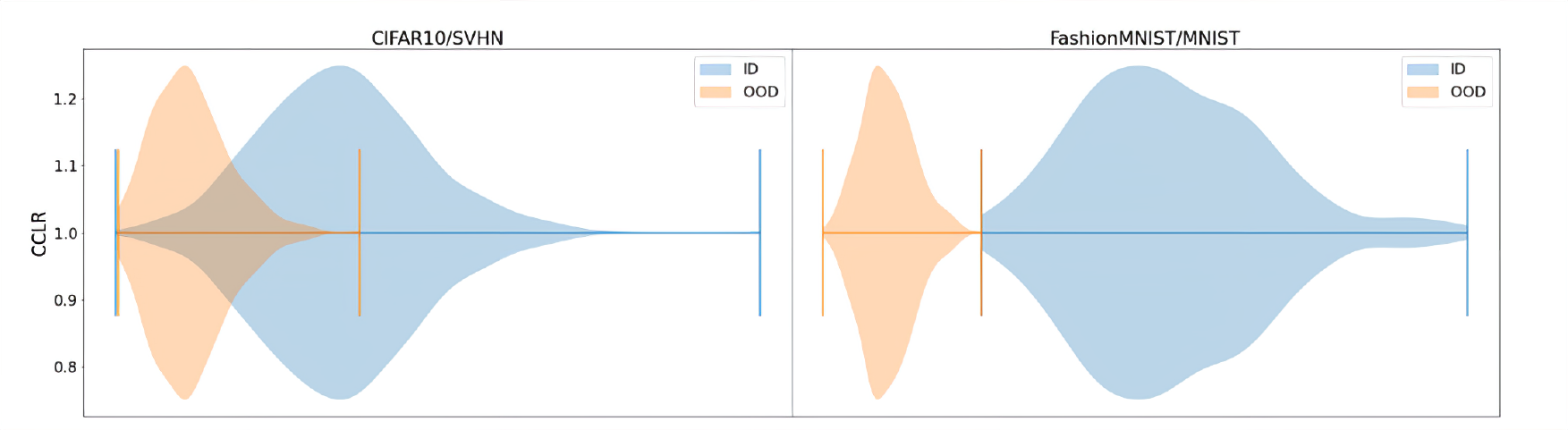}
\caption{\small{Violin plots displaying histograms of the CCLR score for both CIFAR10/SVHN (\textbf{Left}) and FashionMNIST/MNIST (\textbf{Right}) dataset pairs. For CIFAR10/SVHN violin plot, the CCLR scores were calculated using $k/T = 1/2$. For FashionMNIST/MNIST, CCLR scores were calculated using $k/T = 1/5$. Both of the selected $k/T$-values gave the highest AUROC scores from Table 1 for each respective dataset pair.}}
\end{figure}
\begin{table}[h!]
\centering
\scalebox{0.75}{
\begin{tabular}{lll}
\hline
Method                                                                                        & \multicolumn{2}{c}{Dataset Pair}   \\ \hline
-                                                                                             & FashionMNIST/MNIST & CIFAR10/SVHN \\ \hline
\multicolumn{3}{c}{Likelihood-based}                                                                                               \\ \hline
\multicolumn{1}{l|}{WAIC (model ensemble) \cite{choi2018waic}}               & 0.766               & \textbf{1.000}         \\
\multicolumn{1}{l|}{Likelihood Ratio \cite{ren2019likelihood}}               & 0.997               & 0.912        \\
\multicolumn{1}{l|}{Likelihood Regret \cite{xiao2020likelihood}}             & 0.988               & 0.875        \\
\multicolumn{1}{l|}{BIVA with LLR \cite{havtorn2021hierarchical}}               & 0.984               & 0.891        \\
\multicolumn{1}{l|}{MSMA KD Tree \cite{mahmood2020multiscale}}               & 0.693               & 0.991        \\
\multicolumn{1}{l|}{Probabilistic Auto-Encoder \cite{bohm2020probabilistic}} & 0.997               & -            \\
\multicolumn{1}{l|}{S using PCNN and FLIF \cite{serra2019input}}             & 0.967               & 0.929        \\
\multicolumn{1}{l|}{S using Glow and FLIF \cite{serra2019input}}             & 0.998               & 0.950        \\
\multicolumn{1}{l|}{DoSE using VAE \cite{morningstar2021density}}             & 0.998               & -        \\
\multicolumn{1}{l|}{DoSE using GLOW \cite{morningstar2021density}}             & -               & 0.973        \\
\multicolumn{1}{l|}{Local Auto-Regressive Model \cite{zhang2021out}}         & \textbf{1.000}                & 0.969        \\
\multicolumn{1}{l|}{INCP-VAE \cite{ren2019likelihood}}         & \textbf{1.000}                & \textbf{1.000}        \\
\multicolumn{1}{l|}{CCLR using DDPM (Ours)}                                           & \textbf{1.000}                     & 0.964              \\ \hline
\multicolumn{3}{c}{Reconstruction-based}                                                                                           \\ \hline
\multicolumn{1}{l|}{Mahalanobis* \cite{denouden2018improving}}                & 0.986               & 0.991        \\
\multicolumn{1}{l|}{DDPM + SSIM  \cite{graham2022denoising}}                 & 0.974               & 0.979        \\
\multicolumn{1}{l|}{Diffuion-Based Neighborhood \cite{liuout}}                          & -               & 0.950  \\
\multicolumn{1}{l|}{LMD \cite{liu2023unsupervised}}                          & \textbf{0.992}               & \textbf{0.992}      
\end{tabular}}
\caption{\small{AUROC scores for FashionMNIST/MNIST and CIFAR10/SVHN dataset pairs for OOD detection with generative models for both likelihood-based and reconstruction-based methods. \textit{* implies knowledge of OOD labels.}}}

\end{table}
\vspace{-6pt}
In Table 1, we present AUROC results for dataset pairs for RGB and single-channel image dataset comparisons. For each dataset pair, we present AUROC scores using just the ELBO as an OOD score, and for CCLR scores using a range of $k/T$ values. For the FashionMNIST/MNIST dataset pair, the ELBO objective for DDPMs is shown to be a poor estimator for OOD detection. The CCLR for various $k/T$ values proves to be a very strong OOD detection score for single-channel dataset pairs. All $k/T$-values that were $<1/3$ lead to a $1.00$ AUROC score, or the datasets were completely separated using this score up to 3sf (See Fig. 3). For CIFAR10/SVHN, the ELBO objective was a surprisingly good OOD estimator of $0.941$, and is within $0.006$ of the lowest CCLR score. The CCLR proved to be a strong estimator with peak AUROC scores of $0.964$ at $k/T = 1/2$ (See Fig. 3).  \

Our results for FashionMNIST/MNIST are on par with the state-of-the-art results of the Local Auto-Regressive Model \cite{zhang2021out} and INCP-VAE \cite{ran2022detecting} and outperform all current DDPM-based approaches for this dataset pair (See Table 2). For CIFAR10/SVHN our results are competitive with the other likelihood-based methods. However, on this dataset pair, CCLR is outperformed by the reconstruction-based methods using DDPMs.  
\vspace{-15pt}
\section{Conclusions \& Future Work}
In this work, we propose a likelihood ratio for OOD detection, CCLR, with DDPMs. We show that likelihood-based OOD detection can be extended to diffusion models by leveraging the fact that DDPMs, like other likelihood-based generative models, are affected by the input sample complexity. For future work, there have been improvements that have been proposed for DDPMs that could benefit our OOD detection method. These include architecture improvements for the noise prediction network \cite{hoogeboom2023simple}, new image transforms \cite{salimans2022progressive, bansal2022cold} and reformulation of the DDPM objective and noising schedule \cite{kingma2021variational,kingma2023understanding}. Our work uses a vanilla DDPM to act as a proof of concept for likelihood-based OOD detection with DDPMs, and it is likely that performance gains could be made by implementing these improvements. 

\section{Acknowledgements}
This work was supported by Innovative Physics Ltd, the UKRI Centre for Doctoral Training in Accountable, Responsible and Transparent AI (grant number EP/S023437/1) and the UKRI CAMERA project (EP/T022523/1). The authors would like to thank Jessica Nicholson for her help in editing this paper.

\bibliography{egbib}

\begin{thebibliography}{39}
\providecommand{\natexlab}[1]{#1}
\providecommand{\url}[1]{\texttt{#1}}
\expandafter\ifx\csname urlstyle\endcsname\relax
  \providecommand{\doi}[1]{doi: #1}\else
  \providecommand{\doi}{doi: \begingroup \urlstyle{rm}\Url}\fi

\bibitem[Bansal et~al.(2022)Bansal, Borgnia, Chu, Li, Kazemi, Huang, Goldblum,
  Geiping, and Goldstein]{bansal2022cold}
Arpit Bansal, Eitan Borgnia, Hong-Min Chu, Jie~S Li, Hamid Kazemi, Furong
  Huang, Micah Goldblum, Jonas Geiping, and Tom Goldstein.
\newblock Cold diffusion: Inverting arbitrary image transforms without noise.
\newblock \emph{arXiv preprint arXiv:2208.09392}, 2022.

\bibitem[Bishop(1994)]{bishop1994novelty}
Christopher~M Bishop.
\newblock Novelty detection and neural network validation.
\newblock \emph{IEE Proceedings-Vision, Image and Signal processing},
  141\penalty0 (4):\penalty0 217--222, 1994.

\bibitem[B{\"o}hm and Seljak(2020)]{bohm2020probabilistic}
Vanessa B{\"o}hm and Uro{\v{s}} Seljak.
\newblock Probabilistic auto-encoder.
\newblock \emph{arXiv preprint arXiv:2006.05479}, 2020.

\bibitem[Chen et~al.(2020)Chen, Zhang, Zen, Weiss, Norouzi, and
  Chan]{chen2020wavegrad}
Nanxin Chen, Yu~Zhang, Heiga Zen, Ron~J Weiss, Mohammad Norouzi, and William
  Chan.
\newblock Wavegrad: Estimating gradients for waveform generation.
\newblock \emph{arXiv preprint arXiv:2009.00713}, 2020.

\bibitem[Choi et~al.(2018)Choi, Jang, and Alemi]{choi2018waic}
Hyunsun Choi, Eric Jang, and Alexander~A Alemi.
\newblock Waic, but why? generative ensembles for robust anomaly detection.
\newblock \emph{arXiv preprint arXiv:1810.01392}, 2018.

\bibitem[Denouden et~al.(2018)Denouden, Salay, Czarnecki, Abdelzad, Phan, and
  Vernekar]{denouden2018improving}
Taylor Denouden, Rick Salay, Krzysztof Czarnecki, Vahdat Abdelzad, Buu Phan,
  and Sachin Vernekar.
\newblock Improving reconstruction autoencoder out-of-distribution detection
  with mahalanobis distance.
\newblock \emph{arXiv preprint arXiv:1812.02765}, 2018.

\bibitem[Graham et~al.(2022)Graham, Pinaya, Tudosiu, Nachev, Ourselin, and
  Cardoso]{graham2022denoising}
Mark~S Graham, Walter~HL Pinaya, Petru-Daniel Tudosiu, Parashkev Nachev,
  Sebastien Ourselin, and M~Jorge Cardoso.
\newblock Denoising diffusion models for out-of-distribution detection.
\newblock \emph{arXiv preprint arXiv:2211.07740}, 2022.

\bibitem[Havtorn et~al.(2021)Havtorn, Frellsen, Hauberg, and
  Maal{\o}e]{havtorn2021hierarchical}
Jakob~D Havtorn, Jes Frellsen, S{\o}ren Hauberg, and Lars Maal{\o}e.
\newblock Hierarchical vaes know what they don’t know.
\newblock In \emph{International Conference on Machine Learning}, pages
  4117--4128. PMLR, 2021.

\bibitem[Hendrycks et~al.(2018)Hendrycks, Mazeika, and
  Dietterich]{hendrycks2018deep}
Dan Hendrycks, Mantas Mazeika, and Thomas Dietterich.
\newblock Deep anomaly detection with outlier exposure.
\newblock \emph{arXiv preprint arXiv:1812.04606}, 2018.

\bibitem[Ho et~al.(2020)Ho, Jain, and Abbeel]{ho2020denoising}
Jonathan Ho, Ajay Jain, and Pieter Abbeel.
\newblock Denoising diffusion probabilistic models.
\newblock \emph{Advances in Neural Information Processing Systems},
  33:\penalty0 6840--6851, 2020.

\bibitem[Hoogeboom et~al.(2023)Hoogeboom, Heek, and
  Salimans]{hoogeboom2023simple}
Emiel Hoogeboom, Jonathan Heek, and Tim Salimans.
\newblock simple diffusion: End-to-end diffusion for high resolution images.
\newblock \emph{arXiv preprint arXiv:2301.11093}, 2023.

\bibitem[Kingma et~al.(2021)Kingma, Salimans, Poole, and
  Ho]{kingma2021variational}
Diederik Kingma, Tim Salimans, Ben Poole, and Jonathan Ho.
\newblock Variational diffusion models.
\newblock \emph{Advances in neural information processing systems},
  34:\penalty0 21696--21707, 2021.

\bibitem[Kingma and Gao(2023)]{kingma2023understanding}
Diederik~P Kingma and Ruiqi Gao.
\newblock Understanding the diffusion objective as a weighted integral of
  elbos.
\newblock \emph{arXiv e-prints}, pages arXiv--2303, 2023.

\bibitem[Kingma and Welling(2013)]{kingma2013auto}
Diederik~P Kingma and Max Welling.
\newblock Auto-encoding variational bayes.
\newblock \emph{arXiv preprint arXiv:1312.6114}, 2013.

\bibitem[Kingma and Dhariwal(2018)]{kingma2018glow}
Durk~P Kingma and Prafulla Dhariwal.
\newblock Glow: Generative flow with invertible 1x1 convolutions.
\newblock \emph{Advances in neural information processing systems}, 31, 2018.

\bibitem[Krizhevsky et~al.(2009)Krizhevsky, Hinton,
  et~al.]{krizhevsky2009learning}
Alex Krizhevsky, Geoffrey Hinton, et~al.
\newblock Learning multiple layers of features from tiny images.
\newblock 2009.

\bibitem[LeCun et~al.(1998)LeCun, Bottou, Bengio, and
  Haffner]{lecun1998gradient}
Yann LeCun, L{\'e}on Bottou, Yoshua Bengio, and Patrick Haffner.
\newblock Gradient-based learning applied to document recognition.
\newblock \emph{Proceedings of the IEEE}, 86\penalty0 (11):\penalty0
  2278--2324, 1998.

\bibitem[Liu et~al.()Liu, Ren, Cheng, and Zhao]{liuout}
Luping Liu, Yi~Ren, Xize Cheng, and Zhou Zhao.
\newblock Out-of-distribution detection with diffusion-based neighborhood.

\bibitem[Liu et~al.(2023)Liu, Zhou, Wang, and Weinberger]{liu2023unsupervised}
Zhenzhen Liu, Jin~Peng Zhou, Yufan Wang, and Kilian~Q Weinberger.
\newblock Unsupervised out-of-distribution detection with diffusion inpainting.
\newblock \emph{arXiv preprint arXiv:2302.10326}, 2023.

\bibitem[Mahmood et~al.(2020)Mahmood, Oliva, and Styner]{mahmood2020multiscale}
Ahsan Mahmood, Junier Oliva, and Martin Styner.
\newblock Multiscale score matching for out-of-distribution detection.
\newblock \emph{arXiv preprint arXiv:2010.13132}, 2020.

\bibitem[Morningstar et~al.(2021)Morningstar, Ham, Gallagher, Lakshminarayanan,
  Alemi, and Dillon]{morningstar2021density}
Warren Morningstar, Cusuh Ham, Andrew Gallagher, Balaji Lakshminarayanan, Alex
  Alemi, and Joshua Dillon.
\newblock Density of states estimation for out of distribution detection.
\newblock In \emph{International Conference on Artificial Intelligence and
  Statistics}, pages 3232--3240. PMLR, 2021.

\bibitem[Nalisnick et~al.(2018)Nalisnick, Matsukawa, Teh, Gorur, and
  Lakshminarayanan]{nalisnick2018deep}
Eric Nalisnick, Akihiro Matsukawa, Yee~Whye Teh, Dilan Gorur, and Balaji
  Lakshminarayanan.
\newblock Do deep generative models know what they don't know?
\newblock \emph{arXiv preprint arXiv:1810.09136}, 2018.

\bibitem[Netzer et~al.(2011)Netzer, Wang, Coates, Bissacco, Wu, and
  Ng]{netzer2011reading}
Yuval Netzer, Tao Wang, Adam Coates, Alessandro Bissacco, Bo~Wu, and Andrew~Y
  Ng.
\newblock Reading digits in natural images with unsupervised feature learning.
\newblock 2011.

\bibitem[Nichol and Dhariwal(2021)]{nichol2021improved}
Alexander~Quinn Nichol and Prafulla Dhariwal.
\newblock Improved denoising diffusion probabilistic models.
\newblock In \emph{International Conference on Machine Learning}, pages
  8162--8171. PMLR, 2021.

\bibitem[Prince(2023)]{prince2023understanding}
Simon~J.D. Prince.
\newblock \emph{Understanding Deep Learning}.
\newblock MIT Press, 2023.
\newblock URL \url{https://udlbook.github.io/udlbook/}.

\bibitem[Ran et~al.(2022)Ran, Xu, Mei, Xu, and Liu]{ran2022detecting}
Xuming Ran, Mingkun Xu, Lingrui Mei, Qi~Xu, and Quanying Liu.
\newblock Detecting out-of-distribution samples via variational auto-encoder
  with reliable uncertainty estimation.
\newblock \emph{Neural Networks}, 145:\penalty0 199--208, 2022.

\bibitem[Ren et~al.(2019)Ren, Liu, Fertig, Snoek, Poplin, Depristo, Dillon, and
  Lakshminarayanan]{ren2019likelihood}
Jie Ren, Peter~J Liu, Emily Fertig, Jasper Snoek, Ryan Poplin, Mark Depristo,
  Joshua Dillon, and Balaji Lakshminarayanan.
\newblock Likelihood ratios for out-of-distribution detection.
\newblock \emph{Advances in neural information processing systems}, 32, 2019.

\bibitem[Saharia et~al.(2022{\natexlab{a}})Saharia, Chan, Chang, Lee, Ho,
  Salimans, Fleet, and Norouzi]{saharia2022palette}
Chitwan Saharia, William Chan, Huiwen Chang, Chris Lee, Jonathan Ho, Tim
  Salimans, David Fleet, and Mohammad Norouzi.
\newblock Palette: Image-to-image diffusion models.
\newblock In \emph{ACM SIGGRAPH 2022 Conference Proceedings}, pages 1--10,
  2022{\natexlab{a}}.

\bibitem[Saharia et~al.(2022{\natexlab{b}})Saharia, Ho, Chan, Salimans, Fleet,
  and Norouzi]{saharia2022image}
Chitwan Saharia, Jonathan Ho, William Chan, Tim Salimans, David~J Fleet, and
  Mohammad Norouzi.
\newblock Image super-resolution via iterative refinement.
\newblock \emph{IEEE Transactions on Pattern Analysis and Machine
  Intelligence}, 2022{\natexlab{b}}.

\bibitem[Salimans and Ho(2022)]{salimans2022progressive}
Tim Salimans and Jonathan Ho.
\newblock Progressive distillation for fast sampling of diffusion models.
\newblock \emph{arXiv preprint arXiv:2202.00512}, 2022.

\bibitem[Salimans et~al.(2017)Salimans, Karpathy, Chen, and
  Kingma]{salimans2017pixelcnn++}
Tim Salimans, Andrej Karpathy, Xi~Chen, and Diederik~P Kingma.
\newblock Pixelcnn++: Improving the pixelcnn with discretized logistic mixture
  likelihood and other modifications.
\newblock \emph{arXiv preprint arXiv:1701.05517}, 2017.

\bibitem[Serr{\`a} et~al.(2019)Serr{\`a}, {\'A}lvarez, G{\'o}mez, Slizovskaia,
  N{\'u}{\~n}ez, and Luque]{serra2019input}
Joan Serr{\`a}, David {\'A}lvarez, Vicen{\c{c}} G{\'o}mez, Olga Slizovskaia,
  Jos{\'e}~F N{\'u}{\~n}ez, and Jordi Luque.
\newblock Input complexity and out-of-distribution detection with
  likelihood-based generative models.
\newblock \emph{arXiv preprint arXiv:1909.11480}, 2019.

\bibitem[Sohl-Dickstein et~al.(2015)Sohl-Dickstein, Weiss, Maheswaranathan, and
  Ganguli]{sohl2015deep}
Jascha Sohl-Dickstein, Eric Weiss, Niru Maheswaranathan, and Surya Ganguli.
\newblock Deep unsupervised learning using nonequilibrium thermodynamics.
\newblock In \emph{International Conference on Machine Learning}, pages
  2256--2265. PMLR, 2015.

\bibitem[Wang(2020)]{Wang2020}
Phil Wang.
\newblock denoising-diffusion-pytorch.
\newblock \url{https://github.com/lucidrains/denoising-diffusion-pytorch.git},
  2020.

\bibitem[Xiao et~al.(2017)Xiao, Rasul, and Vollgraf]{xiao2017fashion}
Han Xiao, Kashif Rasul, and Roland Vollgraf.
\newblock Fashion-mnist: a novel image dataset for benchmarking machine
  learning algorithms.
\newblock \emph{arXiv preprint arXiv:1708.07747}, 2017.

\bibitem[Xiao et~al.(2020)Xiao, Yan, and Amit]{xiao2020likelihood}
Zhisheng Xiao, Qing Yan, and Yali Amit.
\newblock Likelihood regret: An out-of-distribution detection score for
  variational auto-encoder.
\newblock \emph{Advances in neural information processing systems},
  33:\penalty0 20685--20696, 2020.

\bibitem[Yang et~al.(2022)Yang, Zhang, Song, Hong, Xu, Zhao, Shao, Zhang, Cui,
  and Yang]{yang2022diffusion}
Ling Yang, Zhilong Zhang, Yang Song, Shenda Hong, Runsheng Xu, Yue Zhao,
  Yingxia Shao, Wentao Zhang, Bin Cui, and Ming-Hsuan Yang.
\newblock Diffusion models: A comprehensive survey of methods and applications.
\newblock \emph{arXiv preprint arXiv:2209.00796}, 2022.

\bibitem[Zhang et~al.(2021)Zhang, Zhang, and McDonagh]{zhang2021out}
Mingtian Zhang, Andi Zhang, and Steven McDonagh.
\newblock On the out-of-distribution generalization of probabilistic image
  modelling.
\newblock \emph{Advances in Neural Information Processing Systems},
  34:\penalty0 3811--3823, 2021.

\bibitem[Zhang et~al.(2018)Zhang, Isola, Efros, Shechtman, and
  Wang]{zhang2018unreasonable}
Richard Zhang, Phillip Isola, Alexei~A Efros, Eli Shechtman, and Oliver Wang.
\newblock The unreasonable effectiveness of deep features as a perceptual
  metric.
\newblock In \emph{Proceedings of the IEEE conference on computer vision and
  pattern recognition}, pages 586--595, 2018.

\end{thebibliography}
\end{document}